\documentclass[10pt,twocolumn,letterpaper]{article}

\usepackage{times}
\usepackage{epsfig}
\usepackage{graphicx}
\usepackage{amsmath}
\usepackage{amssymb}

\usepackage[pagenumbers]{cvpr} 

 \usepackage{graphicx}
 \usepackage{amsmath}
 \usepackage{amssymb}
 \usepackage{booktabs}
\usepackage{float}
\usepackage{makecell}
\usepackage{multirow}
\usepackage{multicol}
 \usepackage{cite}
\usepackage{algpseudocode}
\usepackage{algorithm2e}
\RestyleAlgo{ruled}

\usepackage{caption}
\usepackage{subcaption}
\usepackage{tablefootnote}

\usepackage[pagebackref,breaklinks,colorlinks]{hyperref}

\begin{document}

\begin{sloppypar}

\author{
\text{Yusuf Artan, Bensu Alkan Semiz} \\
Image and Video Processing Group, Havelsan Incorporation, Ankara, Turkey
}


\title{\ Fusion of Minutia Cylinder Codes and Minutia Patch Embeddings for Latent Fingerprint Recognition }

\maketitle
\thispagestyle{empty}

\begin{abstract}

Latent fingerprints are one of the most widely used forensic evidence by law enforcement agencies. However, latent recognition performance is far from the exemplary performance of sensor fingerprint recognition due to deformations and artifacts within these images. In this study, we propose a fusion based local matching approach towards latent fingerprint recognition. Recent latent recognition studies typically relied on local descriptor generation methods, in which either handcrafted minutiae features or deep neural network features are extracted around a minutia of interest, in the latent recognition process. Proposed approach would integrate these handcrafted features with a recently proposed deep neural network embedding features in a multistage fusion approach to significantly improve latent recognition results. Effectiveness of the proposed approach has been shown on several public and private data sets. As demonstrated in our experimental results, proposed method improves rank-1 identification accuracy by considerably for real-world datasets when compared to either the single usage of these features or existing state-of-the-art methods in the literature. 

\end{abstract}


\section{Introduction}

\indent Biometrics technology is widely used in the person identification or verification tasks by utilizing physical or behavioral characteristics of an individual such as face, fingerprints, voice, gait or signature to name a few \cite{Jain2004}, \cite{Biometrics1999}. In the past ten years, biometric systems utilizing computer vision technology for fingerprint, iris, palmprint and face recognition tasks are ubiquitously applied in a broad range of fields, e.g., law enforcement, forensics, banking, healthcare and national identity programs \cite{Maltoni2009}. As the most widely adopted biometric solution, automated fingerprint identification system (AFIS) identifies and/or verifies a query fingerprint by comparing it to the reference fingerprints stored in a database. 

Fingerprint recognition task can be broadly divided into two categories; latent and sensor (rolled and slap images) recognition. Sensor based fingerprint recognition methods have achieved exemplary recognition performance thanks to high quality of images and distinctive structures within fingerprints (e.g., bifurcations, ridge endings) \cite{innovatrics}. However, latent fingerprint recognition task has been considered more difficult due to poor quality of images collected from the crime scene \cite{Lee2001}. Moreover, deformation and artifacts within these images also negatively impact the overall recognition performance \cite{Medina2016, Cao2019}. In sensor recognition tasks, local matching methods such as Minutia Cylinder Coding (MCC), minutiae triplets (m-triplets) have shown great success in terms of accuracy, memory and speed performance. These local descriptor based matching methods utilize only spatial coordinates and angle information associated with each minutiae in their analysis \cite{Cappelli2010, Maltoni2009, Medina2011} in which local descriptors extracted within a fixed neighborhood of minutiae are used to measure similarity of images corresponding to rolled and slap fingerprint impressions. 
Even though, these methods are memory efficient and work-well for sensor image datasets, their performance is known to be considerably lower for latent fingerprint recognition \cite{Medina2016, Cao2019, Cao2020}. In a recent study, Medina \textit{et al.} \cite{Medina2016} proposed a deformation tolerant extension of local descriptor algorithms that can handle non-linear distortion present in latent images. This algorithm performs clustering of matching minutiae in an iterative process, in which algorithm finds multiple overlapping clusters of matching minutiae and the best clusters are merged to deal with the nonlinear distortion of the fingerprints. In another recent study, Kılınç \textit{et al.} \cite{Kilinc2022} proposed a mutistage fusion matcher (FM) that would handle the inherent distortion problem by generating mutiple cylinder templates by applying affine transformation and quality thresholding on the extracted minutiae. Many existing legacy commercial-off-the-shelf (COTS) fingerprint recognition systems still rely on some variant of these local matching methods in their matching process.\newline
\indent
In recent years, there have been a rise in the automatic latent fingerprint recognition research using deep neural networks \cite{Cao2019, Cao2020,Ezeobiejesi2018,Tang2018}. Many methods have been proposed towards subtasks of latent fingerprint recognition such as latent image minutia extraction, latent image quality assessment, orientation field estimation and minutia descriptor (embedding) generation tasks \cite{Ezeobiejesi2018, Tang2018}. 
For instance, in a recent study, Cao \textit{et al.} \cite{Cao2020} designed an end-to-end latent fingerprint recognition system that performs automatic segmentation of latent print region, pre-processing, features extraction and matching operations, sequentially. In another study, Tang \textit{et al.} \cite{Tang2018} proposed a unified framework named FingerNet for minutiae extraction task along with orientation field estimation, latent print segmentation, and latent print enhancement. Recently, Ozturk \textit{et al.} \cite{MinNet2022} and Engelsma \textit{et al.} \cite{engelsma2019learning} proposed two distinct deep learning based local path embedding models to represent local ridge flow and spatial/angular minutiae distribution around a fixed neighborhood of a minutia. These earlier works show the great potential of utilizing deep neural networks in the latent fingerprint recognition task.

 
Fusion techniques consolidates information from multiple sources to improve recognition accuracy \cite{Singh2019, Feng2009, Dvornychenko2012, Ross2006, Kong2006, He2010, Madeena2017}. These techniques are known to produce significant performance gains on latent recognition task as shown in earlier studies \cite{Indovina2011,Indovina2012}. Combined information utilized by these systems are likely to be more distinctive compared to the information obtained from a single source. There are various configurations of the biometric fusion techniques such as multi-sensor, multi-algorithm, multi-instance, multi-sample and multi-modal \cite{Paulino2010, Singh2019}. Since the AFIS databases typically contain only a single impression of latent image and a single impression of corresponding sensor image, multi-algorithm based fusion is more appropriate for latent recognition task. In terms of the levels of the fusion, feature-level, score-level and rank-level fusion are commonly used for latent recognition task \cite{Veeramachaneni2008, Ross2006, Madeena2017}. For example, Baig \textit{et al.} \cite{Waqas2018} and Benhammadi \textit{et al.} \cite{Benhammadi2007} proposed a feature level fusion method that could utilize both the minutiae and texture features in the matching process. Jain \textit{et al.} \cite{Jain2008} and several other studies noted that score level fusion achieves better performance improvements compared to feature and image fusion approaches \cite{Feng2009, Singh2019, He2010}. 

In this study, we aim to integrate handcrafted minutia features (e.g., minutia cylinder codes (MCC)) with local patch embeddings of a novel deep learning architecture to exploit benefits of both deep neural networks (DNNs) and traditional domain knowledge inspired features. This fusion approach allows us to augment existing features of the legacy AFIS system with discriminative embedding vectors of deep network architectures. To this end, we aggregate minutia pairings obtained from Fusion Matcher (an \cite{Kilinc2022} and recently proposed MinNet model to improve latent recognition as explained in section \ref{sec:Methods}. Note that fusion of MCC and DNN embeddings have not been investigated in earlier studies. As shown in experimental results, proposed method improves performance significantly compared to single usage of these features and existing matching algorithms.  

Remainder of the paper is organized as follows: Section \ref{sec:relatedwork} briefly presents a review of related works in the literature. Section \ref{sec:Methods} discusses the details of the proposed approach. Section \ref{sec:experiments} describes the details of the datasets used in this study and performance analysis of the proposed method. Finally, conclusion are presented in section \ref{sec:conclusion}.

\begin{figure*}[t]
    \centering
    \includegraphics[width=\textwidth]{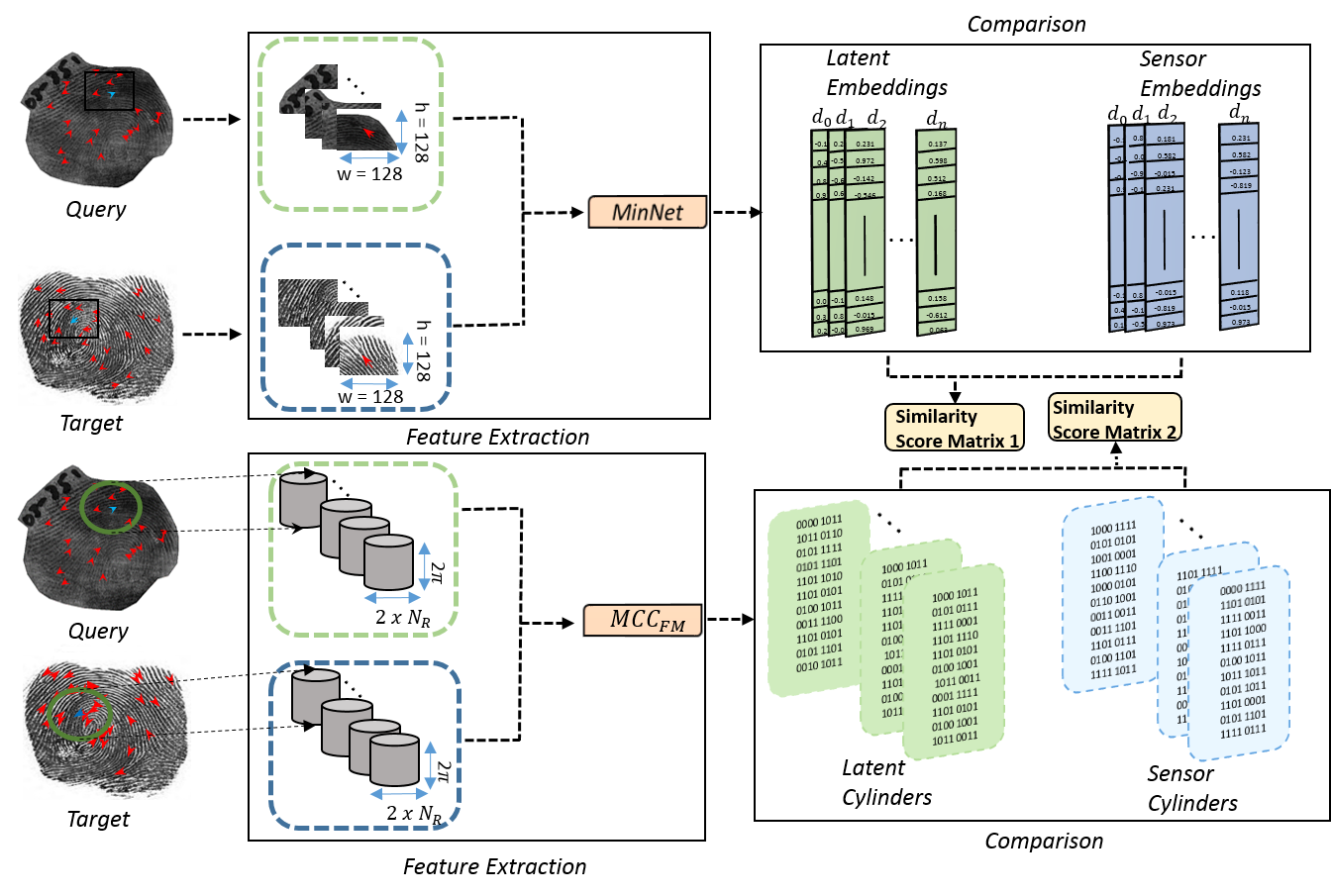}
    \caption{This figure illustrates an overview of the proposed feature level fusion approach towards latent fingerprint recognition task. MCC cylinder vectors and MinNet embedding vectors are generated around each minutiae (shown in red dots) for both query (latent) and target (sensor) images. Next, 2 separate similarity score matrices are generated corresponding to MCC cylinders and MinNet embeddings, and top matching minutiae pairs are selected from these similarity score matrices. Finally, we concatenate these matching minutiae pairs for 2 different descriptors and generate a match score using Local Similarity Sort with Relaxation (LSS-R) algorithm. }
    \label{fig:my_label}
\end{figure*}

\section{Related Work}\label{sec:relatedwork}

\textbf{Minutia Cylinder Codes (MCC):} is a fixed-length orientation invariant descriptor that is extracted from a local fixed neighborhood around each minutia of the minutiae template $T$ \cite{Cappelli2010}. This representation encodes spatial and angular relationships between a minutia and its-fixed radius neighborhood in the form of a cylinder whose based and height are related to the spatial and directional information, respectively. Let a given ISO/IEC 197974-2 \cite{ISO2005} minutia template $T=\{ m_{1}, m_{2}, \cdots, m_{N}\}$ consist of a set of minutiae extracted from a fingerprint; each minutiae m is a triplet $m_{i}=\{ x_{i}, y_{i}, \theta_{i}\}$ where $(x_{i}, y_{i})$ is the spatial position and $\theta_{i}$ is the minutia direction (in the range $[0,2\pi]$). Height of the cylinder is divided into $N_{D}$ sections, where each section corresponds to a directional difference in the range $[-\pi,\pi[$. Cylinder base are discretized into cells ($N_{S}$ is the number of cells), and for each cell a numerical value is calculated by accumulating contributions from minutiae in a neighborhood of the projection of the cell center onto the cylinder base. Contribution of each minutia $m_{t}$ to a cell of the cylinder (corresponding to a given minutia $m$) depends on the spatial and directional information of the neighboring minutiae \cite{Cappelli2010}.
Once cylinders are generated for templates corresponding to rolled and latent templates. Matching minutiae pairs are determined from the similarity score matrix generated by measuring similarity of cylinders.

\textbf{MinNet Minutia Embeddings:} Minutiae Patch Embedding Network (MinNet) is a recently introduced deep learning model that generates an embedding vector to represent local image patch cropped around a minutia \cite{MinNet2022}. Unlike earlier models, proposed local descriptor model learns to represent local ridge flow and spatial/angular minutiae distribution around a fixed neighborhood of a minutiae simultaneously through a novel cost function. Before the patch extraction step, fingerprint image are enhanced to eliminate noise and to enhance local ridge flow. Once the enhancement operation completes, fingerprint region is segmented from the image to remove unnecessary background. Therefore, we extract segmentation masks, enhancement maps and minutiae using FingerNet \cite{Tang2018} algorithm that is trained for latent fingerprint minutiae extraction task. These operations bring latent fingerprints and sensor fingerprints into the same domain as shown in \cite{MinNet2022}. MinNet utilizes MobileNetV3-large based backbone model to extract features that are used to generate embedding vectors that would also preserve spatial and angular distributions of neighboring minutiae. While the Minutia Segmentation branch of the MinNet model aims to reconstruct neighboring minutiae positional and angular information, Descriptor Generation branch creates a patch embedding vector to represent ridge flow information of the patch. In the training process, additive angular margin loss (AAM) term \cite{deng2019arcface} is used as a loss function for descriptor generation and mean squared error (MSE) is used as the loss function for Minutia Segmentation model. In the end, each minutia patch is represented as a $1\times256$ floating point embedding vector. 

\section{Methods}\label{sec:Methods}
This study proposes a novel feature level fusion approach towards latent fingerprint recognition task using minutia cylinder codes and MinNet embeddings. To this end, minutia cylinder codes and embedding vectors are generated for each minutiae of sensor (target) and latent (query) images as shown in Fig. \ref{fig:my_label}. Next, we measure the local similarity between latent minutia embedding/code ($v_{i}$) and sensor minutia embedding/code ($v_{j}$) using cosine similarity measure, $s( v_{i}, v_{j} ) = \frac{ v_{i}^{T}\cdot v_{j} }{\|v_{i}\| \|v_{j}\|}$. Therefore, for a given sensor (target) and latent (query) minutia embedding templates $A= \{a_{1},a_{2}, \dots, a_{M} \}$ and $B= \{b_{1},b_{2}, \dots, b_{N} \}$, respectively;\newline
\indent
$\bullet$ $s(a,b)$ denote the local similarity between minutia $a\in A$ and $b\in B$, with $s(\cdot): A \times B$ $\xrightarrow{}[-1,1]$,\newline
\indent
$\bullet$ $\Gamma \in [0,1]^{n_{A} \times n_{B}}$ denote the similarity matrix corresponding to embedding templates $A$ and $B$ containing the local similarity between minutia embeddings with $\Gamma[r,c]=s(a_{r},b_{c})$.\newline
\indent
When comparing two minutiae embedding templates (corresponding to latent image and sensor image), we deduce a set of matching minutiae pairs by performing Local Similarity Assigment (LSA) on the similarity score matrix in which we determine a set of $n_{R}$ minutiae pairs that would attain the highest similarity score values without considering the same minutia more than once. After that, local similarity sort with relaxation (LSS-R) method is utilized to measure global compatibility of the initial match pairs as discussed next. 

\SetKwComment{Comment}{/* }{ */}
\begin{algorithm}[ht]
\caption{Proposed Algorithm}\label{alg:main}
\KwData{Minutiae Template A, Minutiae Template B, Patch Embeddings A, Patch Embeddings B, MCC Cylinders A, MCC Cylinders B ,  $\delta_{theta}$ }
\KwOut{Score = MatchScore} 
\vspace{1em}
$T_{A}  \gets  \textrm{Minutiae Template A}$ 

$T_{B} \gets  \textrm{Minutiae Template B}$ 

$MinNet_{A} \gets  \textrm{Patch Embeddings A} $ 

$MinNet_{B} \gets  \textrm{Patch Embeddings B} $ 

$MCC_{A} \gets \textrm{MCC Cylinders A}$ 

$MCC_{B} \gets \textrm{MCC Cylinders B}$

$S_{MCC} \gets \textrm{SimScore}( MCC_{A}, MCC_{B}, T_{A}, T_{B} )$ 

$S_{MinNet}  \gets \textrm{SimScore}( MinNet_{A} , MinNet_{B})$ 

$n_{R}=\min(12, \min(len(A),len(B)))$

$n_{p}=\min(8, \min(len(A),len(B)))$

\Comment{LSA is used to find pairing minutiae.}
$\lambda_{MCC_{pairs}} = LSA( S_{MCC}, n_{R} )$  

$\lambda_{MinNet_{pairs}} = LSA( S_{MinNet}, n_{R} )$ 

$\lambda_{Pairs}  \gets  \{ \lambda_{MCC_{pairs}} \cup \lambda_{MinNet_{pairs}}$ \} 

\Comment{LSS-R used to compute similarity score using pairing minutiae and corresponding local minutiae similarity scores.}
MatchScore = LSS-R( $\lambda_{Pairs}, n_{p}, Score_{Pairs}$ )

\textrm{\textbf{def SimScore( A, B, $T_{A}, T_{B}$ ):}} \Comment{Similarity Score Generator} 

\If{ (argc$<$3 )}{
$\theta_{A}, \theta_{B} = 0, 0 $
}
s = zeros(len(A),len(B)) 
\For{i = \textrm{1:len(A)}}{ 
\indent          $v_{i} = A_{i}$\\
\indent           $\theta_{A} = T_{A}[i].Angle$
\indent        \For{i = \textrm{1:len(B)}}{
\indent              $v_{j} = B_{j}$ 
\indent              $\theta_{B} = T_{B}[i].Angle$ 
\indent   \If{ $|\theta_{A}-\theta_{B}| \leq \delta_{theta}$ }{

\indent   $s_{i,j} =\frac{ v_{i}^{T}\cdot v_{j} }{\|v_{i}\| \|v_{j}\|}$ 
        }
 }
}
return s
\end{algorithm}

\textbf{LSS-R:} is used to measure the global compatibility between minutiae pairings. It is actually a penalization method for geometrically dissimilar minutiae pairs, in which uncompatability of a pair corresponds to the higher the penalization of the similarity value obtained from similarity score matrix. Let $\gamma_{t}^{0} = \Gamma[r_{t},c_{t}]$ be the initial similarity of pair $t$, and the similarity at iteration $i$ of the relaxation process is given in Eq. \ref{eq:LSSR},
 \begin{equation}\label{eq:LSSR}
      \gamma_{t}^{i} = w_{R}\cdot \gamma_{t}^{i-1} + (1-w_{R})\cdot \frac{ \sum_{k=1, k\neq t}^{n_{R}}\rho(t,k)\cdot \gamma_{k}^{i-1} }{n_{R}-1}
 \end{equation}
where $w_{R} \in [0,1]$ is linear weighting factor and $p(t,k)$ is a measure of compatibility between the pairs of minutiae from template A and template B. $p(t,k)$ compatibility is based on the similarity between minutiae spatial distances, directional differences and radial angles \cite{Cappelli2010} as shown in Eq. \ref{eq1}. 
\begin{equation}\label{eq1}
\begin{split}
p(t,k) & = \prod_{i=1}^{3}\frac{1}{( 1 + \exp{-\tau_{i}(d_{i}-\mu_{i})}  )} \\
 d_{1} & = | d_{S}(a_{r_{t}},a_{r_{k}}) - d_{S}(b_{r_{t}},b_{r_{k}}) | \\
 d_{2} & = | d\Phi( d_{\theta}(a_{r_{t}},a_{r_{k}}),d_{\theta}(b_{r_{t}},b_{r_{k}}) ) | \\
 d_{3} & = | d\Phi( d_{R}(a_{r_{t}},a_{r_{k}}),d_{R}(b_{r_{t}},b_{r_{k}}) ) |
\end{split}
\end{equation}
where $d_{S}, d_{\theta} $ and $d_{R}$ corresponds to euclidean distance, angular distance and radial distance among matching minutiae pairs as explained in \cite{Cappelli2010}. $d\Phi(\theta_{1},\theta_{2})$ denotes the difference between angles $\theta_{1}$ and $\theta_{2}$ as shown in Eq. \ref{eq:dPhi} . After $n_{rel}$ iterations of the relaxation process are executed on all the $n_{R}$ pairs, top $n_{p}$ minutiae pairings are selected to produce the final matching score by summing up similarity score values of these $n_{p}$ pairings.
\begin{equation}\label{eq:dPhi}
d\Phi(\theta_{1},\theta_{2}) = \mathrm{min}( |\theta_{1}-\theta_{2}|, 2\pi-|\theta_{1}-\theta_{2}| )
\end{equation}
\indent In this study, two separate minutiae template similarity score matrices ($S_{MCC}$ and $S_{MinNet}$) are generated by comparing the query (latent) minutiae cylinders/embeddings and the target (sensor) minutiae cylinders/embeddings as shown in Fig. \ref{fig:my_label}. Next, we aggregated matching minutia pairs derived from MCC and MinNet similarity matrices and passed these minutia pairs along with their similarity scores to the LSS-R algorithm to determine a global similarity score. Algorithm-1 presents the details of the proposed approach where $n_{R}$ minutiae pairs with top similarity values from $S_{MCC}$ and $S_{MinNet}$ are merged to generate the final match pair indices. In the experiments, we refer to the proposed feature-level approach as $P_{feature}$, and we set $n_{p}=\min(8, \min(N,M))$.
 

In addition to the feature level fusion approach ($P_{feature}$), we have also investigated the performance of score level and rank-level fusion. In the score level fusion, a weighted  similarity score $S_{score}$ is computed as the weighted sum of $S_{MCC}$ and $S_{MinNet}$ matrices as shown in Equation \ref{eq:score_fusion}.
\begin{equation}\label{eq:score_fusion}
    S_{score} = w_{1} \cdot S_{MCC} +  w_{2} \cdot S_{MinNet}
\end{equation}
, where $w_{1}\in [0,1]$ and $w_{2}\in [0,1]$ are the weights corresponding to MCC and MinNet features, respectively. Next, $n_{R}$ matching minutiae pairs are selected from $S_{score}$ matrix using LSA algorithm mentioned earlier. Then, LSS-R process is applied on these minutiae pairs to compute final match score. In our experiments, we referred to score-level fusion results as $P_{score}$. \newline
\indent
Finally, in the rank based fusion of approach, individual MCC and MinNet based match results are sorted to determine rank information. Fused rank value is declared as the minimum rank (highest rank) \cite{TinKam1994} achieved for each query sample. Rank-1 identification is declared if MCC and MinNet matchers find the target at rank-1. We referred to rank-level fusion results as $P_{rank}$ in section \ref{sec:experiments}.

\section{Experiments} \label{sec:experiments}

\subsection{Datasets} \label{sec:Datasets}
 To evaluate the effectiveness of the proposed fusion approach, we have utilized several public and private datasets as described next. 


\subsubsection{\textbf{EGM Test Dataset (Private):}} 
This dataset is obtained from General Directorate of Police (EGM) in Turkey. 
It contains a total of 11132 fingers that have 2 impression per finger (one latent and its corresponding rolled finger). These rolled images are collected using inkpad and optical scanners. The first row in Fig. \ref{fig:EGM_datasets} illustrates sample image pairs from EGM dataset.

\subsubsection{\textbf{JGK Test Dataset (Private):}} 
\begin{figure}[htp]
    \centering
    \begin{tabular}{ c c c c }
      \includegraphics[width=1.8cm]{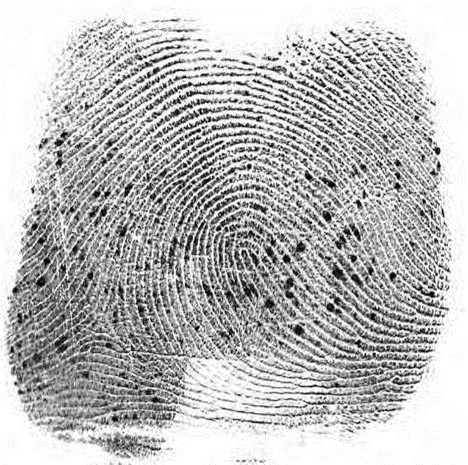} &  \includegraphics[width=1.7cm]{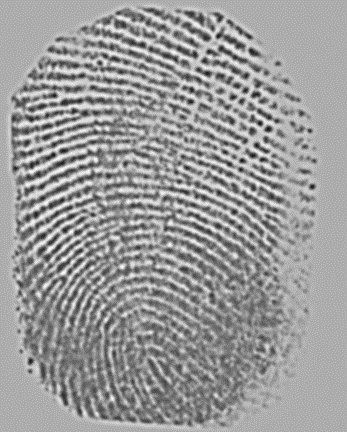} &
      \includegraphics[width=1.9cm]{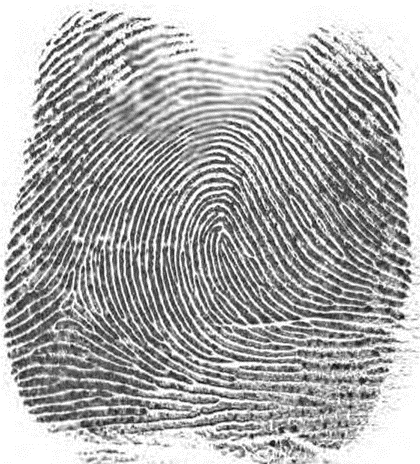} & \includegraphics[width=1.7cm]{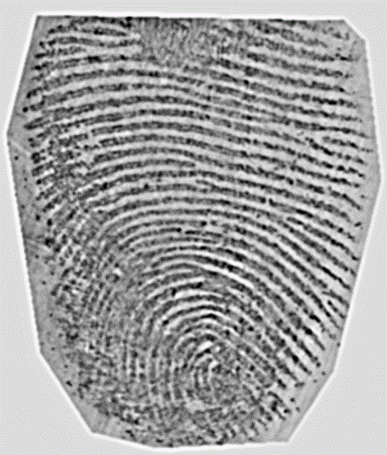}\\
      \includegraphics[width=1.5cm]{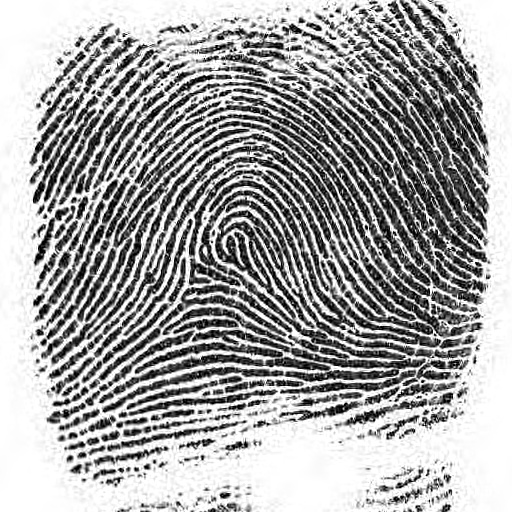} &  \includegraphics[width=1.8cm]{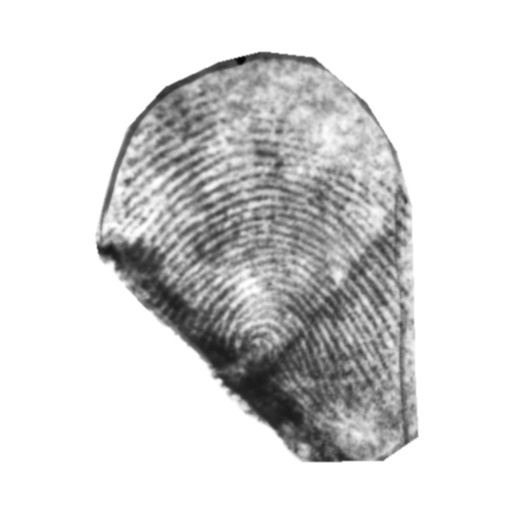} &
      \includegraphics[width=1.5cm]{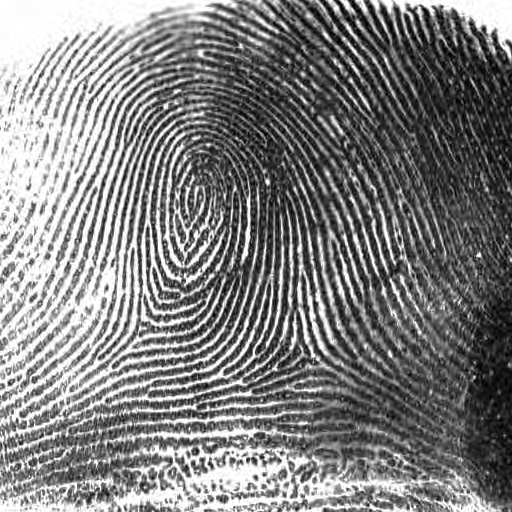} & \includegraphics[width=1.5cm]{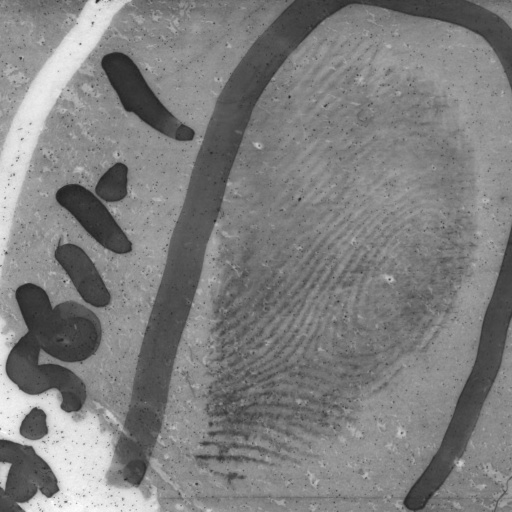}\\
      (a) & (b) & (c) & (d) \\
     \end{tabular}
    \caption{Sample images from private EGM Test Dataset (first-row) and JGK Test Dataset (second row). Pairs of rolled (left) and latent (right) images from EGM and dataset; (a)-(b) and (c)-(d). }
    \label{fig:EGM_datasets}
 \end{figure}

This dataset is obtained from the databases of the Gendarmerie General Command (JGK) in Turkey. This dataset contains a total of 4042 fingers that have 2 impression per finger (as one latent and one rolled finger). The second row in Fig. \ref{fig:EGM_datasets} presents sample image pairs from JGK dataset.

\subsubsection{\textbf{FVC-Latent Test Dataset (Public):}} 
\begin{figure}
\begin{center}
    
\begin{subfigure}{0.24\columnwidth}
\includegraphics[width=\textwidth]{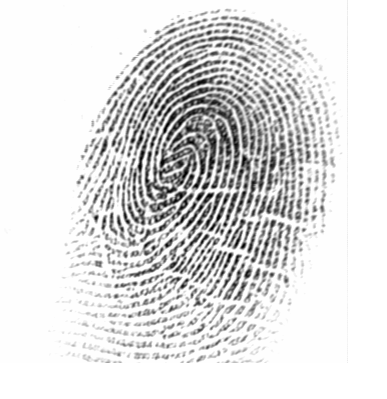}
\caption{}
\end{subfigure}
\begin{subfigure}{0.24\columnwidth}
\includegraphics[width=\textwidth]{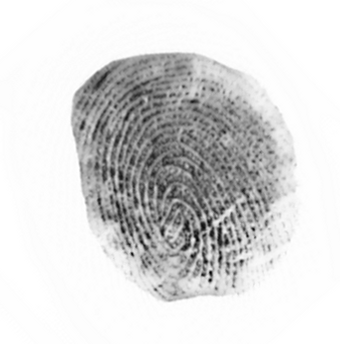}
\caption{}
\end{subfigure}
\begin{subfigure}{0.24\columnwidth}
\includegraphics[width=\textwidth]{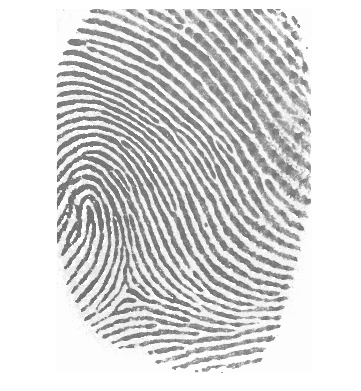}
\caption{}
\end{subfigure}
\begin{subfigure}{0.24\columnwidth}
\includegraphics[width=\textwidth]{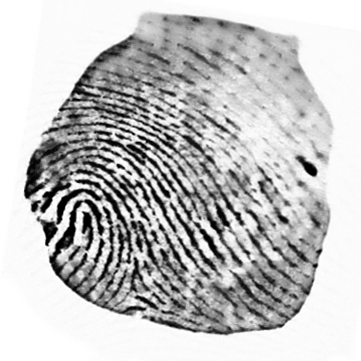}
\caption{}
\end{subfigure}
\caption {FVC Latent Test Dataset Examples a-c) Slap Fingerprint b-d) Fake Latent Fingerprint }
\label{fig:FVC_Latent_samples}
\end{center}
\end{figure}

FVC is a multi-dataset collected using various sensor technologies e.g., optical, thermal and capacitive \cite{Maio2000,Maio2002_v2,Maio2004}. A recent study \cite{MinNet2022} extended this dataset for latent fingerprint recognition task by emulating latent like images corresponding to some images within this dataset. These latent like images are generated using a deep learning based general adversial network model (GAN).  To this end,  \cite{MinNet2022} trained a CycleGan model and applied random distortion, random rotation and selection operations, sequentially, on the sensor images to emulate latent images. In this study, we utilized FVC-Latent test set comprising of 316 image pairs released at the link\footnote{\url{https://github.com/FingerGeneration/Generated_Latent_Fingerprint_Dataset}}. Fig. \ref{fig:FVC_Latent_samples} presents several samples from this emulated dataset. Since FVC latent considerably small test data set, we augmented FVC test data set with 10K rolled images to enlarge reference data set.

\vspace{-0.2cm}
\subsection{Results}\label{sec:Results}
In this section, we report performance results of the proposed technique for datasets listed in section \ref{sec:Datasets}. In addition to the feature based fusion results, we also discuss performances of the rank-level, score-level based fusion results and existing methods presented in  MCC \cite{Cappelli2010}, MinNet \cite{MinNet2022}, and \cite{Cao2020}. When reproducing results for the study of Cao \textit{et al.} \cite{Cao2020}, we utilized SDK provided by the authors. In terms of the MCC algorithm \cite{Cappelli2010}, we utilized our in-house developed MCC variant that has some modifications to the original MCC algorithm, which also performs better than author provided SDK on these datasets. Furthermore, in our comparisons, we have also utilized a multistep fusion extension of MCC algorithm that utilizes scale, rotation and quality attributes of minutiae in the cylinder generation and matching phases \cite{Kilinc2022}. This algorithm will be referred to as $\mathrm{MCC_{FM}}$ in our experiments.
Throughout our experiments, we utilized minutiaes extracted by FingerNet \cite{Tang2018} algorithm, except for the method of \cite{Cao2020} which utilizes an autoencoder model to extract its own minutiae, when evaluating the performance of MCC \cite{Cappelli2010}, MinNet \cite{MinNet2022} and the proposed method. For $\tau$ and $\mu$ parameters in Eq. \ref{eq1}, we utilize the values listed in Table \ref{tab:tau_mu} and $w_{R}$ parameter is set to $0.5$ in Eq. \ref{eq:LSSR}.

\begin{table}[htbp]
\begin{tabular}{cccccccc}
\Xhline{1.3pt}
Params & $\mu_{1}$  & $\tau_{1}$  & $\mu_{2}$ & $\tau_{2}$ & $\mu_{3}$ & $\tau_{3}$    \\
\hline
Values & 0.0416 & -30 & 0.7853 & -9 & 0.2094 & -16.8 \\
\hline
\end{tabular}
\caption{ $\tau_{i}$ and $\mu_{i}$ parameters used in Eq. \ref{eq1}. }
\label{tab:tau_mu}
\end{table}

\begin{figure*}[t]
\centering
\begin{tabular}{ c c c c }
 & 
\includegraphics[width=5.1cm]{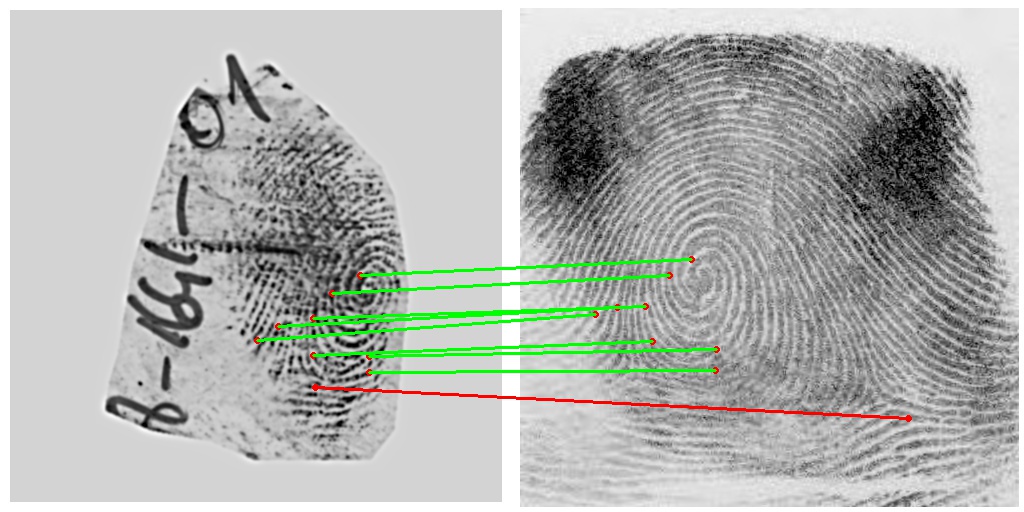} &  \includegraphics[width=5.1cm]{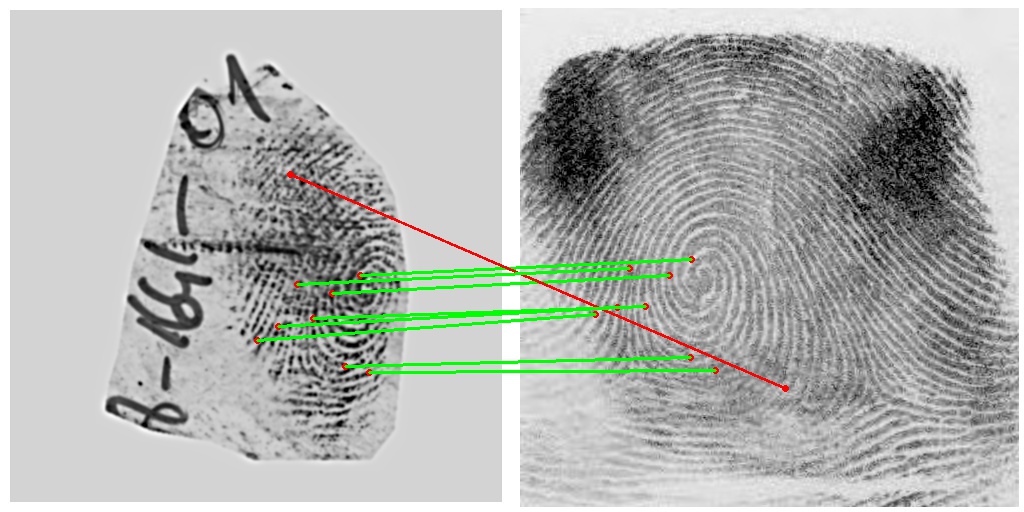} &
\includegraphics[width=5.1cm]{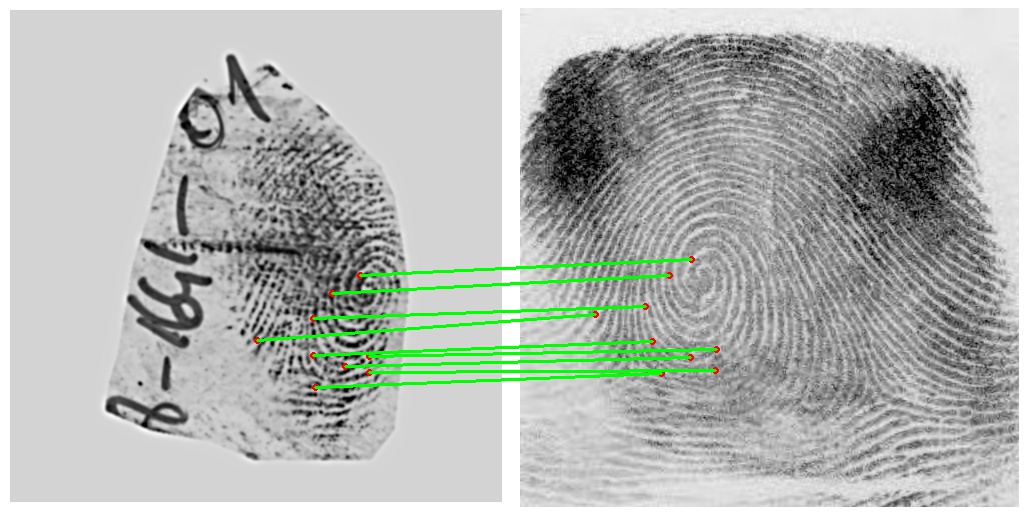} \\
\hline
\thead{Method}& \thead{MinNet} & \thead{$MCC_{FM}$} & \thead{ $P_{feature}$} \\
\hline
\thead{Rank} &\thead{1} & \thead{2} & \thead{1} \\
\hline
\end{tabular}
\caption{ This figure presents a sample latent-rolled pair image from EGM Test Dataset. A visual illustration of the best matching 8 minutiae pair correspondences for (left) MinNet \cite{MinNet2022}, (middle) $MCC_{FM}$ and (right) $P_{feature}$. While MinNet \cite{MinNet2022} and $MCC_{FM}$ is able to match in rank-1 and rank-2, respectively, there exists a false matched minutiae correspondence shown in red line.}
\label{fig:simScores}
\end{figure*}

\subsubsection{\textbf{EGM Test Dataset Results:}}

\begin{figure}[ht]
    \centering
     \includegraphics[width=\linewidth-0.5cm]{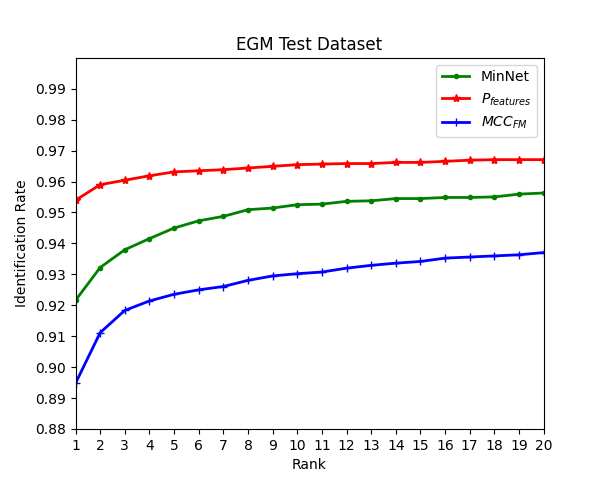} 
    \caption{ Cumulative Match Characteristic (CMC) curves on EGM Test Dataset for top-3 performing methods listed in Table \ref{tab:EGM_results}.  }
    \label{fig:EGM_results}
\end{figure}
 
For this dataset, we evaluated the recognition performance of proposed method for each latent image by comparing it to 5560 sensor images in our database. Table \ref{tab:EGM_results} presents a comparison of rank-1, rank-5 and rank-10 accuracy values of our proposed method along with methods of \cite{Cao2020} and \cite{Cappelli2010}. Fig. \ref{fig:EGM_results} shows the cumulative match characteristic curve (CMC) for top-3 performing methods listed in Table \ref{tab:EGM_results}. 
Proposed feature-level fusion technique outperforms earlier methods considerably (in terms of rank-1 identification rate) on this data set. MCC and MinNet algoritms may return false matched minutiae correspondences (for a genuine match) between a latent and its rolled mate, hence a lower match score at the output of LSS-R algorithm. Proposed method, on the other hand, is able improve true minutiae correspondences and weaken false minutiae correspondences by fusing highest minutiae correspondences obtained from MCC and MinNet as shown in Fig. \ref{fig:simScores}. Examination of the match results show that some of the highest 8 minutiae correspondences are false for both MinNet and MCC algorithms but proposed method correctly assigns the highest minutiae correspondences. In terms of the score-level fusion, $w_{1}=w_{2}=0.5$ yields the best result reported in Table \ref{tab:EGM_results}. In LSS-R algorithm, compatible minutiae pairs increase match score of genuine match, hence, this affect global similarity match score, positively. We should also note that even though rank-5 and rank-10 performance of the $P_{rank}$ is slightly higher then $P_{feature}$, in real world scenarios rank-1 is considered more important than the others.

 \begin{table} [h] 
    \begin{center}
         \begin{tabular}{ c |  c | c | c } 
        \Xhline{1.3pt}
          \multirow{3}{*}{\parbox{1.5cm}{\thead{Methods}}} & \multicolumn{3}{c}{ \thead{EGM Test Dataset}} \\
         \cline{2-4} & \thead{Rank-1} & \thead{Rank-5} &  \thead{Rank-10}   \\ [0.5ex] 
        \Xhline{1.0pt}
          \thead{\cite{Cao2020}} & \thead{85.88 } & \thead{ 88.91} & \thead{89.92}  \\
         \hline
         \thead{MCC \cite{Cappelli2010}} & \thead{80.59} & \thead{84.98} & \thead{86.66}  \\
         \hline
         \thead{$\mathrm{MCC_{FM}}$ \cite{Kilinc2022}} & \thead{89.47} & \thead{92.35} & \thead{93.02}  \\
         \hline
         \thead{MinNet \cite{MinNet2022}} &  \thead{92.39} & \thead{94.71}&\thead{95.30} \\
         \hline
         \thead{ $P_{rank}$ } & \thead{{86.37}} & \thead{\textbf{98.72}} & \thead{\textbf{98.83}}    \\
         \hline
         \thead{ $P_{score}$ } & \thead{{95.13}} & \thead{{96.15}} & \thead{96.49}  \\
         \hline
         \thead{ $P_{feature}$ } & \thead{\textbf{95.39}} & \thead{96.31} & \thead{96.54}    \\
        \Xhline{1.3pt}
        \end{tabular}
       \caption{Rank-1, rank-5 and rank-10 identification accuracy values for EGM Test dataset using the proposed method and earlier methods in the literature. Proposed feature-level fusion approach yields the best rank-1 identification rate.}
       \label{tab:EGM_results}
    \end{center}
\end{table}

\subsubsection{\textbf{JGK Test Dataset Results:}}

 \begin{table}[h] 
    \begin{center}
         \begin{tabular}{ c |  c | c | c } 
        \Xhline{1.3pt}
          \multirow{3}{*}{\parbox{1.5cm}{\thead{Methods}}} & \multicolumn{3}{c}{ \thead{JGK Test Dataset}} \\
         \cline{2-4} & \thead{Rank-1} & \thead{Rank-5} &  \thead{Rank-10}   \\ [0.5ex] 
        \Xhline{1.0pt}
          \thead{\cite{Cao2020}} & \thead{81.79} & \thead{86.98} & \thead{87.82}  \\
         \hline
         \thead{MCC \cite{Cappelli2010}} & \thead{75.25} & \thead{81.29} & \thead{83.17}  \\
         \hline
         \thead{$\mathrm{MCC_{FM}}$ \cite{Kilinc2022}} & \thead{87.87} & \thead{91.39} & \thead{92.57}  \\
         \hline
         \thead{MinNet \cite{MinNet2022}} & \thead{82.73} & \thead{88.22} & \thead{89.31} \\
         \hline
         \thead{ $P_{rank}$ } & \thead{81.69} & \thead{\textbf{94.31}} & \thead{\textbf{94.40}}    \\
         \hline
         \thead{ $P_{score}$ } & \thead{88.17} & \thead{91.83} & \thead{92.47}    \\
         \hline
         \thead{ $P_{feature}$ } & \thead{\textbf{90.95}} & \thead{93.32} & \thead{93.62}    \\
        \Xhline{1.3pt}
        \end{tabular}
       \caption{Rank-1, rank-5 and rank-10 identification accuracy values for JGK Test dataset using the proposed method and earlier methods in the literature. Proposed feature-level fusion approach yields the best rank-1 identification rate. }
       \label{tab:JGK_results}
    \end{center}
\end{table}

\begin{figure}[ht]
\centering %
\includegraphics[width=\linewidth-0.5cm]{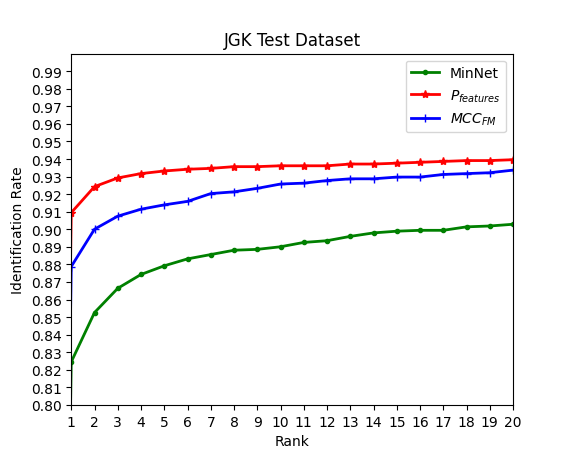}  
\caption{ Cumulative Match Characteristic (CMC) curves on JGK Test Dataset for top-3 performing methods listed in Table \ref{tab:JGK_results}.  }
\label{fig:exp_JGK_results}
\end{figure}

For JGK test dataset, we generated a match score for each latent image by comparing it to 2021 sensor images in our database. Table \ref{tab:JGK_results} presents a comparison of rank-1, rank-5 and rank-10 accuracy values of our proposed method along with methods of \cite{Cao2020}, \cite{Kilinc2022} and \cite{Cappelli2010}. Similar to the EGM Test data set results, proposed fusion rank-1 performance of the proposed method is significantly higher than that of the other methods. Similar to EGM Test dataset results, score-level fusion again yields best results for $w_{1}=w_{2}=0.5$, but its performance is slightly inferior to feature-level fusion. Fig. \ref{fig:exp_JGK_results} shows CMC curves for methods listed in Table \ref{tab:JGK_results}. 

\subsubsection{\textbf{FVC-Latent Test Dataset Results:}}
\begin{figure}[htbp]
    \centering
    \includegraphics[width=\linewidth-0.5cm]{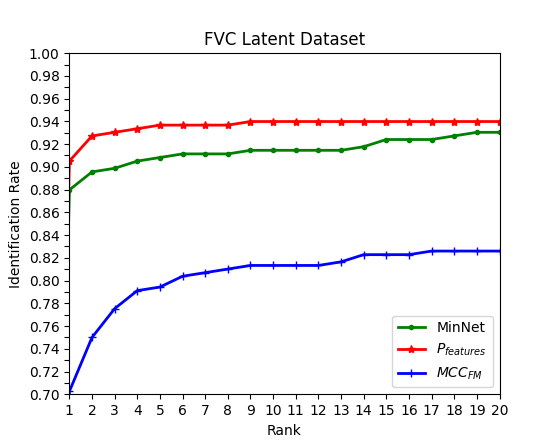} 
    \caption{ Cumulative Match Characteristic (CMC) curves for top-3 performing methods (listed in Table \ref{tab:FVC_results}) on FVC Latent Test dataset.  }
    \label{fig:exp_FVC_results}
 \end{figure}

We performed proposed technique and earlier studies on FVC Latent dataset as well. Table \ref{tab:FVC_results} presents a comparison of rank-1, rank-5 and rank-10 accuracy values of our proposed method along with methods of \cite{Cao2020}, \cite{Medina2016} and \cite{Cappelli2010}. Fig. \ref{fig:exp_FVC_results} shows CMC curves for top-3 methods listed in Table \ref{tab:FVC_results} for FVC-Latent dataset. 

\begin{table}[htbp] 
    \begin{center}
         \begin{tabular}{ c |  c | c | c } 
        \Xhline{1.3pt}
          \multirow{3}{*}{\parbox{1.5cm}{\thead{Methods}}} & \multicolumn{3}{c}{ \thead{FVC Latent}} \\
         \cline{2-4} & \thead{Rank-1} & \thead{Rank-5} &  \thead{Rank-10}   \\ [0.5ex] 
        \Xhline{1.0pt} 
          \thead{\cite{Cao2020}} & \thead{50.15} & \thead{56.46} & \thead{59.62}  \\
         \hline
        \thead{MCC \cite{Cappelli2010}} & \thead{62.37} & \thead{69.77} & \thead{72.34}  \\
         \hline
        \thead{$\mathrm{MCC_{FM}}$ \cite{Kilinc2022}} & \thead{70.25} & \thead{79.45} & \thead{82.34}  \\
         \hline
        \thead{MinNet \cite{MinNet2022}} &  \thead{87.97} & \thead{91.11} & \thead{91.45}    \\
         \hline
         \thead{ $P_{rank}$ } & \thead{{64.63}} & \thead{95.81} & \thead{\textbf{96.46}}    \\
         \hline
         \thead{ $P_{score}$ } & \thead{\textbf{93.98}} & \thead{\textbf{95.88}} & \thead{95.88}    \\  
         \hline
         \thead{$P_{features}$} & \thead{90.50} & \thead{93.67} & \thead{93.98}    \\
        \Xhline{1.3pt}
        \end{tabular}
     \caption{Rank-1, rank-5 and rank-10 identification accuracy values for FVC-Latent Test dataset using the proposed method and earlier methods in the literature. Proposed feature-level fusion approach yields the best rank-1 identification rate.}
     \label{tab:FVC_results}
    \end{center}
\end{table}

 The score-level fusion leads to an additional only 11 matches are retrieved at rank-1 compared to the feature-level fusion. The result may be misleading as it is a considerably small data set compared to JGK and EGM data sets. Feature-level fusion still performs better than the other methods. Similar to the JGK and EGM datasets, the rank-level fusion reaches the highest accuracy score in rank-10. 

\section{Conclusion}\label{sec:conclusion}

In this study, we proposed a feature level fusion approach for latent image recognition task. This approach allows us to combine handcrafted minutia features with local patch embeddings of a state-of-the-art deep learning architecture to exploit benefits of both deep networks and traditional domain knowledge inspired features. This fusion approach allowed us to augment existing features of the legacy system with discriminative embedding vectors of deep network architectures. Proposed approach has been tested on several datasets and has shown to improve performance significantly compared to single usage of these features and existing state-of-the-art matching algorithms.


\end{sloppypar}

\end{document}